\documentclass[runningheads]{llncs}
\usepackage{graphicx}
\usepackage[pagebackref,breaklinks,colorlinks]{hyperref}
\hypersetup{colorlinks, citecolor=blue} 
\usepackage{caption}
\usepackage{float}
\usepackage{subcaption}
\usepackage[table]{xcolor}
\usepackage{multirow}
\usepackage{booktabs}

\begin{document}
\title{Line Graphics Digitization: A Step Towards Full Automation} 

%
%
\author{Omar Moured\inst{1,2} \and
Jiaming Zhang\inst{1}\and
Alina Roitberg\inst{1}\and
Thorsten Schwarz \inst{2}\and
Rainer Stiefelhagen\inst{1,2}}
%
%
%
%
\authorrunning{Omar Moured et al.}
%
\institute{
$^1$CV:HCI lab, Karlsruhe Institute of Technology, Germany.\\
$^2$ACCESS@KIT, Karlsruhe Institute of Technology, Germany.
\email{\{firstname.lastname\}@kit.edu}\\
\url{https://github.com/moured/Document-Graphics-Digitization.git}}



%
\maketitle    
%
\begin{figure}
\includegraphics[width=\textwidth]{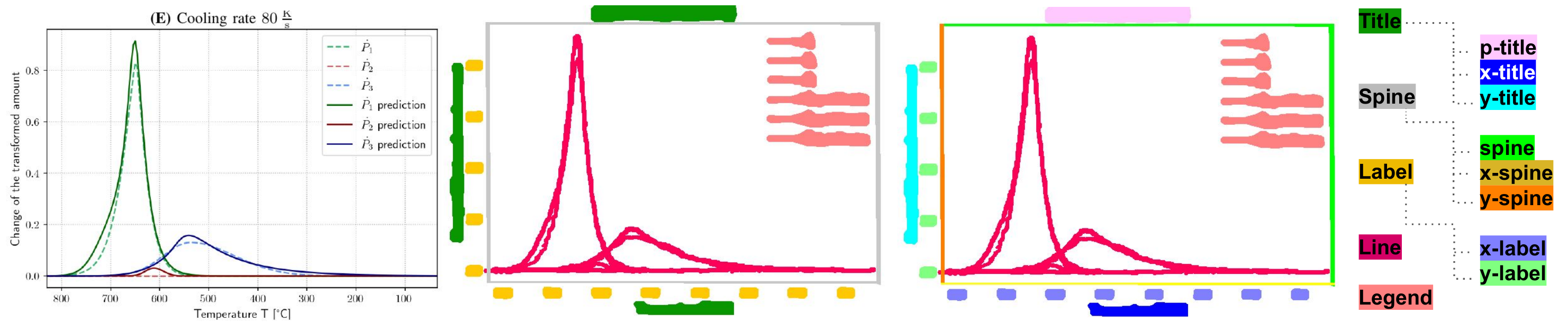}
    \begin{minipage}[t]{.27\textwidth}
        \vskip -3ex
        \subcaption{Input image}\label{fig1_1}
    \end{minipage}%
    \begin{minipage}[t]{.27\textwidth}
        \vskip -3ex
        \subcaption{Coarse classes}\label{fig1_2}
    \end{minipage}%
    \begin{minipage}[t]{.27\textwidth}
        \vskip -3ex
        \subcaption{Fine classes}\label{fig1_3}
    \end{minipage}%
    \begin{minipage}[t]{.19\textwidth}
        \vskip -3ex
        \subcaption{Hierarchy}\label{fig1_4}
    \end{minipage}%
\caption{Hierarchical semantic segmentation on the proposed Line Graphics (LG) benchmark. (a) Each image includes coarse-to-fine two-level semantic segmentation, \textit{i.e.}, (b) coarse 5 classes and (c) 10 fine classes. (d) The coarse-to-fine hierarchy shows the two-level semantic segmentation classes.} \label{cover}
\end{figure} 
%
\begin{abstract}



The digitization of documents allows for wider accessibility and reproducibility. 
While automatic digitization of document layout and text content has been a long-standing focus of research, this problem in regard to graphical elements, such as statistical plots, has been under-explored.
In this paper, we introduce the task of fine-grained visual understanding of mathematical graphics and present the \emph{Line Graphics (LG)} dataset, which includes pixel-wise annotations of $5$ coarse and $10$ fine-grained categories. Our dataset covers $520$ images of mathematical graphics collected from $450$ documents from different disciplines. Our proposed dataset can support two different computer vision tasks, \emph{i.e.}, \emph{semantic segmentation} and \emph{object detection}. To benchmark our LG dataset, we explore $7$ state-of-the-art models.
To foster further research on the digitization of statistical graphs, we will make the dataset, code and models publicly available to the community.

\end{abstract}

\section{Introduction}

With the rapid growth of information available online\footnote{\url{https://www.statista.com/statistics/871513/worldwide-data-created/}}, access to knowledge has never been easier. 
However, as the volume of information continues to grow, there is a need for more efficient ways to extract useful information from documents such as papers and presentation slides. 
This is particularly important for individuals with special needs, such as visually impaired individuals~\cite{keefer2014image}, for whom traditional methods of accessing information may not be feasible.

\begin{figure}
\includegraphics[width=\textwidth]{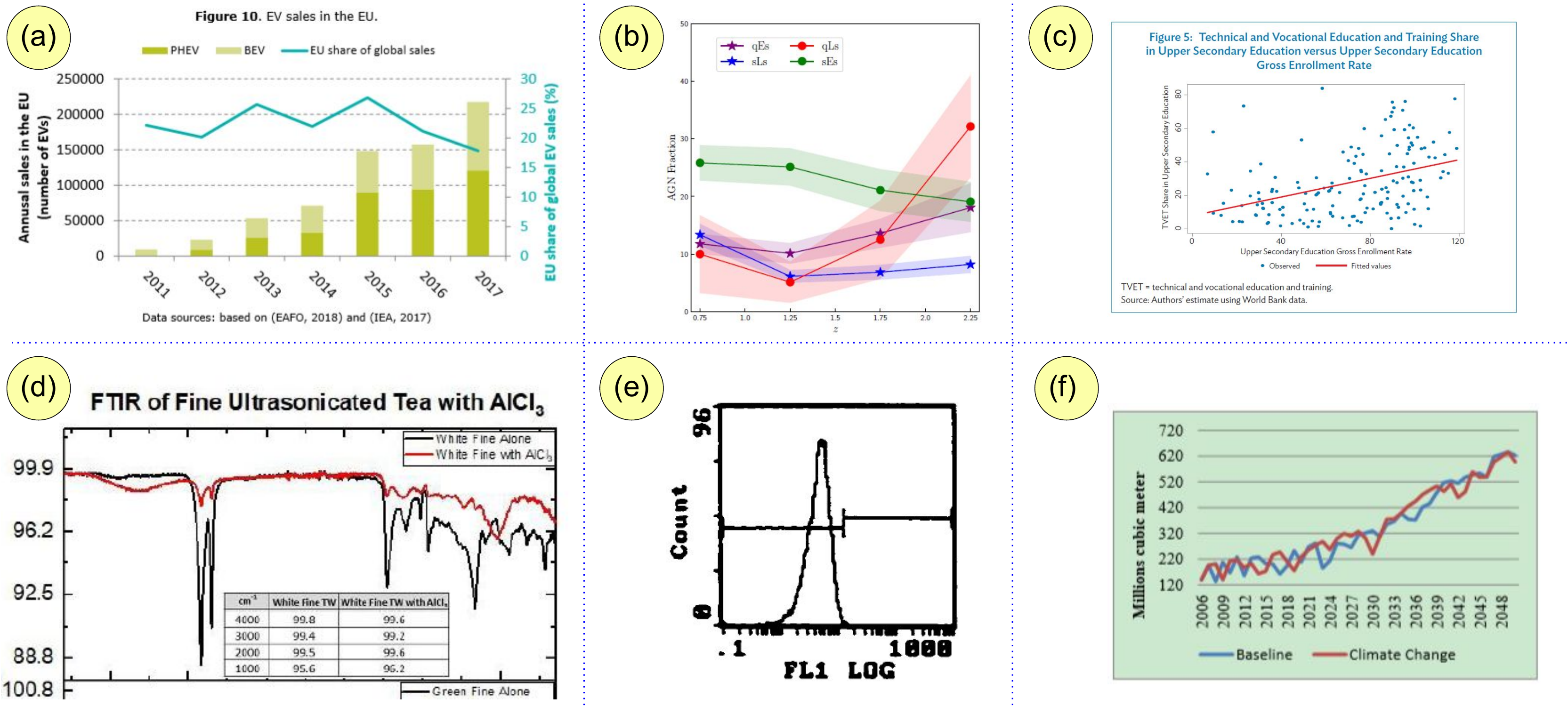}
\caption{Diverse mathematical graphics covered in our Line Graphics (LG) dataset, including 100 bar charts (a), 320 line graphics (b, d-f) and 100 scatter plots (c). These samples pose significant challenges for existing document analysis methods.} 
\label{fig1}
\end{figure}

During courses, graphs are a vital supplement to lecturers' speech as they effectively summarize complex data or visualize mathematical functions.
However, one downside of this medium is the difficulty of automatic information extraction, as  graphs contain very fine-grained elements, such as fine lines, small numbers or axes descriptions, while the traditional document analysis frameworks focus  on coarse structures within complete pages~\cite{clark2016pdffigures,yang2017learning,chen2015page} or slides~\cite{haurilet2019wise,haurilet2019spase}. 
The process of separating distinct regions of a plot and assigning them a semantic meaning at a pixel-level, known as graph segmentation, is an important prerequisite step for graph understanding. One application of using pixel-level data to fully automate the process is to generate an imposed document or 2D refreshable tactile display that can be easily interpreted through touch for people with blindness or visual impairment. Hence, end-to-end full automation of plot digitization could be achieved.  

Presumably, due to the lack of annotated datasets for fine-grained analysis of  plots, the utilization of modern deep semantic segmentation architectures has been rather overlooked in the context of mathematical graphs. 

In this paper, we introduce the task of fine-grained visual understanding of mathematical graphics and present the Line Graphics (LG) dataset, which includes pixel-wise annotations of 10 different categories. 
Our dataset covers 520 images of mathematical graphics collected from 450 documents from different disciplines, such as physics, economics, and engineering. 
Figure \ref{fig1} provides several examples of statistical plots collected in our dataset. 
By providing pixel-wise and bounding box annotations, we enable our dataset to support two different computer vision tasks: \emph{instance, semantic segmentation} and \emph{object detection}.

To benchmark our LG dataset, we explore  7 state-of-the-art models, including efficiency- and accuracy-driven frameworks, (e.g., MobileNetV3~\cite{howard2019searching} and SegNeXt~\cite{guo2022segnext})  with SegNeXt yielding the best results with $67.56\%$ mIoU. 
Our results show that while we have achieved high overall accuracy in our models, the accuracy varies depending on the type of object. 
Specifically, we found that spine-related categories and plot title were the hardest to recognize accurately.
To foster further research on the digitization of statistical graphs, we will make the dataset, code, and models publicly available to the community.

The key findings and contributions of this paper are can be summarized as:
\begin{itemize}
    \item We introduce the task of fine-grained visual understanding of mathematical graphics, aimed at reducing manual user input when digitalizing documents.
    \item We collect and publicly release the Line Graphics (LG) dataset as a benchmark for semantic segmentation and object detection in line graphics. The dataset includes plots from papers and slides from various fields and is annotated with 10 fine-grained classes at both pixel and bounding box levels.
    \item We perform extensive evaluations on 7 state-of-the-art semantic segmentation models, analyzing the impact of factors such as image resolution and category types on the performance. Our findings demonstrate the feasibility of the proposed task, with the top model achieving a mean Intersection over Union 
    of $67.56\%$. However, further advancement is needed in certain categories, such as plot title or spines, as well as for low-resolution data.
\end{itemize}

\section{Related Work}

\subsection{Document Graphics Analysis}
Visual document analysis is a well-studied research area, mostly focusing on text~\cite{long2018textsnake,huang2019icdar2019} and  layout analysis~\cite{chen2015page,chen2017convolutional,clausner2017icdar2017} of complete pages originating from scientific papers~\cite{chintalapati2022dataset,davila2021icpr}, presentation slides~\cite{haurilet2019wise}, magazines~\cite{clausner2017icdar2017}, historical handwritten documents~\cite{chen2015page,chen2017convolutional} or receipts~\cite{huang2019icdar2019}.
In comparison, the research of chart analysis is  more limited, with an 
overview of the existing approaches provided by~\cite{davila2020chart}.
In particular, several learning-based methods have been used for (1) localizing and extracting charts from pages~\cite{clark2016pdffigures,li2019figure,siegel2018extracting}, or (2) harvesting text and tabular data from charts~\cite{choi2019visualizing,liu2019data,methani2020plotqa,davila2021icpr}.
Seweryn et. al.~\cite{seweryn2021will} propose a framework covering chart classification, detection of essential elements, and generation of textual descriptions of four chart types (lines, dot lines, vertical bar plots and horizontal bar plots). 
\cite{chintalapati2022dataset} focuses on exploring the semantic content of alt text in accessibility publications, revealing that the quality of author-written alt text is mixed. The authors also provide a dataset of such alt text to aid the development of tools for better authoring and give recommendations for publishers.
\cite{yoshitake2020program}  developed a program for fully automatic conversions of line plots (png files) into numerical data (CSV files) by using several Deep-NNs.
\cite{bajic2022multi} introduce a fully automated chart data extraction algorithm for circular-shaped and grid-like chart types.
Semantic segmentation of documents~\cite{yang2017learning,amin2001page,drivas1995page,ha1995document,breuel2017robust,haurilet2019spase,haurilet2019wise} has close ties to computer vision, where segmentation model performance has greatly improved with deep learning advancements~\cite{zhang2021transfer,xie2021segformer,guo2022segnext,zhao2017pspnet,chen2018deeplabv3+}. Despite this, research in fine-grained semantic segmentation of mathematical graphs lags due to a lack of annotated examples for training. To address this, our dataset seeks to close the gap and provide a public benchmark for data-driven graph segmentation methods.

\begin{table}[t]
\caption{Overview of the five most related datasets. Our LG dataset for the first time addresses fine-grained semantic segmentation of line graphs.}\label{tab1}
\resizebox{\textwidth}{!}{
\begin{tabular}{l|l|l|l}
\toprule
\textbf{Dataset} & \textbf{Task} & \textbf{Labels Type} & \textbf{Year}  \\ \hline \hline
PDFFigures 2.0 \cite{clark2016pdffigures} & Figure and caption detection& bounding boxes & 2016 \\ 
Poco \textit{et al.} \cite{poco2017reverse} &  Text Pixel Classification and localization & text binary mask and metadata & 2017  \\ 
Dai \textit{et al.} \cite{dai2018chart} &  Classification and text recognition & bounding boxes & 2018  \\ 
DocFigure \cite{jobin2019docfigure}& Figure classification & one hot encoded images  & 2019 \\ 
ICDAR Charts 2019 \cite{davila2019icdar} & Text content extraction and recognition &  bounding boxes & 2019 \\ \hline
\textbf{Ours} & Semantic segmentation and detection & segmentation masks for 10 classes & 2023 \\ 
\bottomrule
\end{tabular}
}
\end{table}

\subsection{Document Graphics Datasets} 
Table \ref{tab1} provides an overview of the five published datasets most related to our benchmark. 
The PDFFigures 2.0 dataset is a random sample of 346 papers from over 200 venues with at least 9 citations collected from Semantic Scholar, covering bounding box annotations for captions, figures, and tables~\cite{clark2016pdffigures}.
The dataset of Poco et al.~\cite{poco2017reverse} comprises automatically generated and manually annotated charts from Quartz news and academic paper.
 The data for each image includes the bounding boxes and transcribed content of all text elements.
 The authors further investigate automatic recovery of visual encodings from chart images using text elements and OCR. They present an end-to-end pipeline that detects text elements, classifies their role and content, and infers encoding specification using a CNN for mark type classification. 
Dai et al.~\cite{dai2018chart} collect a benchmark that covers bar charts collected from the web as well as synthetic charts randomly generated through a script. The authors present Chart Decoder --  a deep learning-based system which automatically extracts textual and numeric information from such charts.
DocFigure~\cite{jobin2019docfigure} is a scientific figure classification dataset consisting of 33,000 annotated figures from 28 categories found in scientific articles. The authors also designed a web-based annotation tool to efficiently categorize a large number of figures. 
The ICDAR 2019 CHART-Infographics competition~\cite{davila2019icdar} aimed to explore automatic chart recognition, which was divided into multiple tasks, including image classification, text detection, text role classification, axis analysis and  plot element detection. A large synthetic training set was provided and systems were evaluated on synthetic charts and real charts from scientific literature.

In comparison to these datasets, \emph{LG} targets \textit{semantic segmentation} of line graphs with $>500$ examples collected from documents of $18$ different disciplines, with manual pixel-level annotations at two levels of granularity and 15 labels in total (5 coarse and 10 fine-grained categories). Our LG dataset aims to establish a public benchmark for data-driven graph segmentation methods and will be made publicly accessible upon publication.



%
\section{\textit{LG} Dataset}
In this paper, we present the first segmentation dataset to analyze line charts and keep pace with the advancements in the AI community. Our dataset contains $520$ mathematical graphics extracted manually from 450 documents. Among these, 7238 human annotated instances. The goal is to facilitate automatic visual understanding of mathematical charts by offering a suitable and challenging benchmark. Next, we provide a comprehensive description of the data collection and annotation process, followed by a thorough analysis of the dataset's features and characteristics.
\begin{figure}[!t]
\centering
\includegraphics[width=\textwidth]{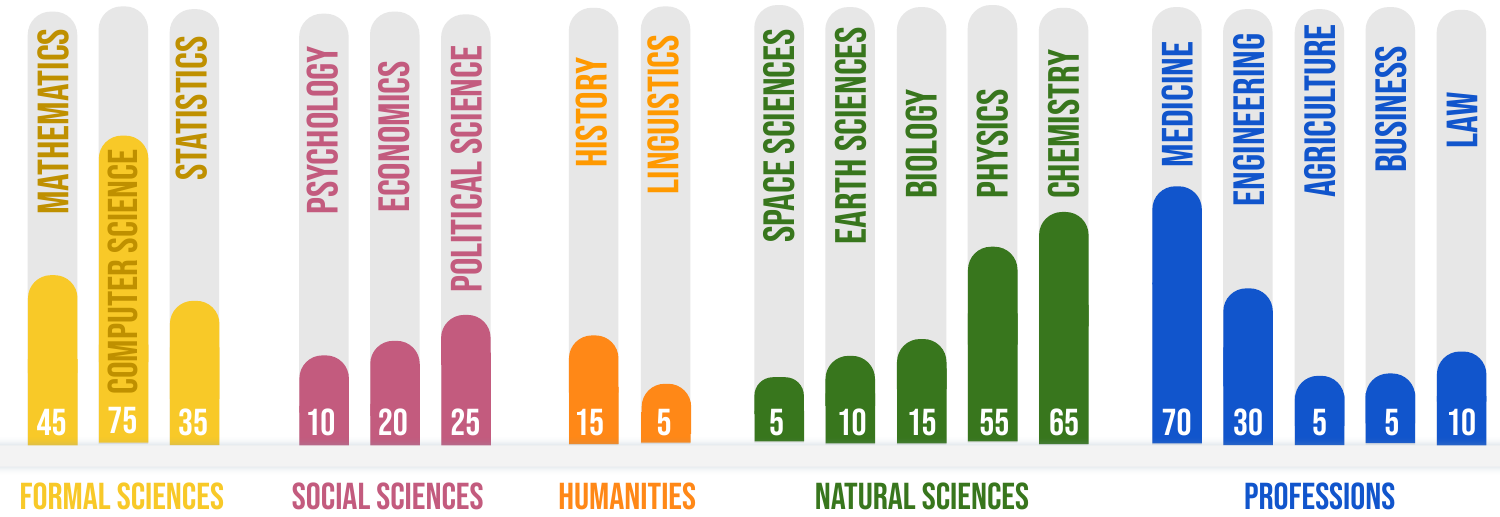}
\caption{Statistical  distribution of documents in our dataset grouped by different disciplines. Our dataset was collected from 18 distinct disciplines from formal-, social and natural sciences as well as humanities and professions.}
\label{doc_disp_stats}
\end{figure}

\subsection{Data Collection and Annotation}
\subsubsection{Classes.} To ensure a comprehensive and robust labelling process, we set out to categorize line chart pixels into 5 coarse and 10 fine-grained classes. The primary focus was on creating fine-grained categories that offer a wide range of variations and challenges for further analysis. This was achieved through a thorough review of charts by three annotators with research experience, who identified the most frequent and critical object types encountered in such charts. Based on this review, as well as an inspection of related work, we arrived at 10 relevant categories. Some of these can be further categorized into three coarse categories, namely, \emph{Title} class (e.g. plot title), \emph{Spine} class (e.g. "spine" with no label data), and the \emph{Label} class (e.g. x-axis labels). As detailed in Table \ref{instances_split}, in this work, we conduct experiments with the $10$ classes, which are\textit{ p-title, x-title, y-title, x-spine, y-spine, spine, x-label, y-label, legend and line}.

\subsubsection{Collection.} In addition to ensuring that the source documents in \emph{LG} dataset are free from intellectual property constraints, we have imposed certain requirements for the documents to adhere to. First, all collected documents should contain at least one complex line chart, regardless of the document type (scanned, digital, slide, etc..) or field. Second, to represent different time periods, both old and new charts were collected. Third, the similarity levels between cropped images were maintained as low as possible to ensure that each image presents a unique and challenging point for analysis. To achieve broad coverage of all fields, documents in \emph{LG} dataset were collected from 5 different disciplines and their top published subcategories as shown in Figure \ref{doc_disp_stats}. The collection process involved a manual search using scientific keywords, and carefully inspecting each document downloaded from sources such as arXiv and Google Scholar. This approach helped ensure a consistent and uniform distribution of documents across all categories.


\begin{figure}[t]
\centering
\includegraphics[width=\textwidth]{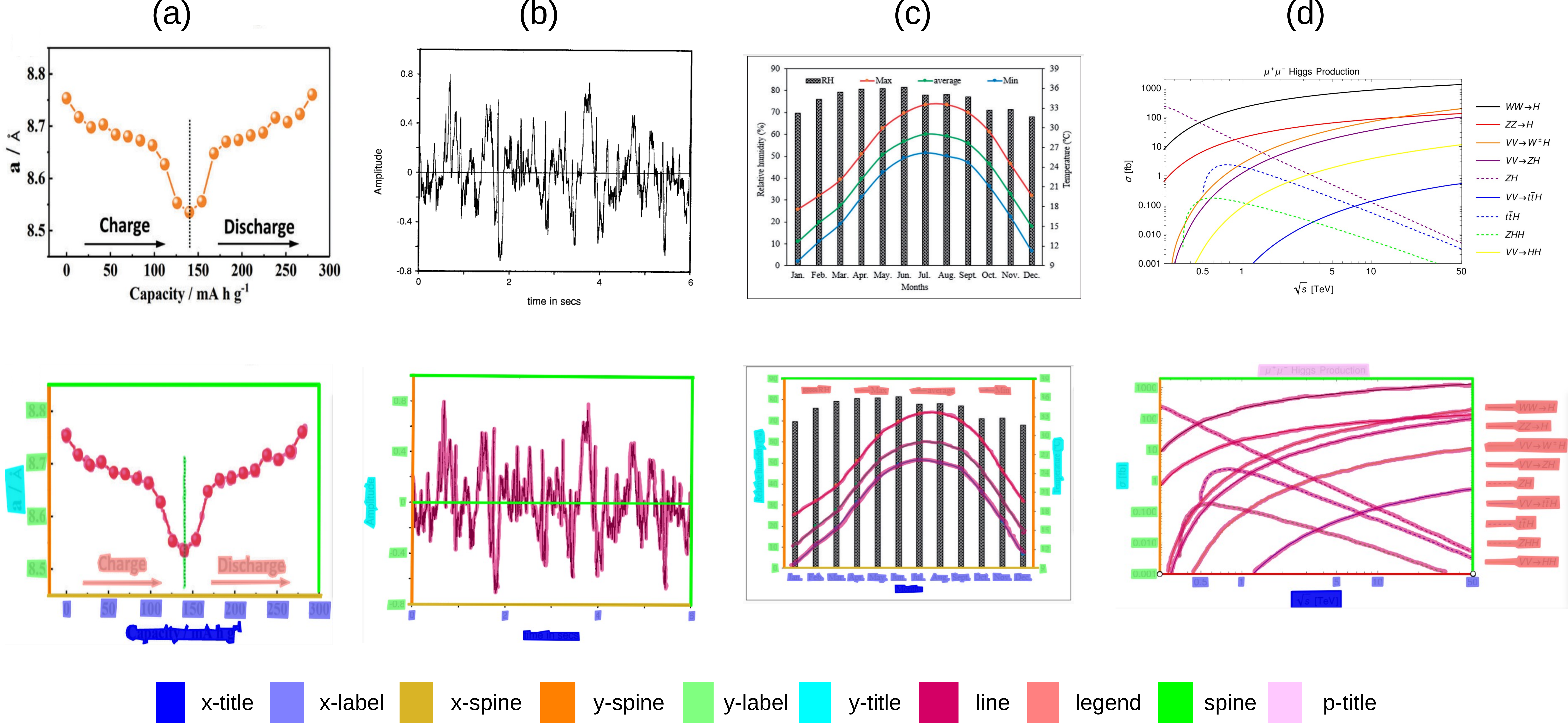}
\caption{Example annotations of our Line Graphics (LG) dataset. From top to bottom are the challenging line graphics and the ground truth with fine-grained annotations of 10  classes, which are complemented by 5 coarse categories. 
}
\label{imgs_with_gt_masks}
\end{figure}

\subsubsection{Annotation.} Fine-grained pixel-level annotations were provided for the 10 chart classes as depicted in Figure \ref{imgs_with_gt_masks}. This level of detail was necessary due to the presence of fine structures in the charts, such as lines. Using bounding boxes alone would not be sufficient as it would result in the incorrect annotation of background pixels as foreground and difficulty in distinguishing between different lines in the plot. Bounding boxes were still provided despite the pixel-level ones, as we believe they may be useful for certain classes such as the text content, except lines, for which a plotting area bounding box category was labelled instead. The annotation process was initiated with the provision of 20 pages of guidelines and 100 sample images to each of the three annotators, resulting in a mean pairwise label agreement of 80\%. Further, annotators were given a batch of images to annotate and each annotation was reviewed by the other two annotators. To facilitate instance-level segmentation, we provide annotations for each instance separately in COCO \textit{JSON} format. For example, each line has a separate ID in the line mask.



\subsection{Dataset Properties} 
Fully automating the mathematical graphics digitization process involves retrieving the metadata and converting it into a machine-readable format, such as a spreadsheet. This requires a thorough understanding of fine-grained elements, such as axes to project the lines and obtain pixel $(x,y)$ entries, axes labels to calibrate retrieved line values from the pixels domain, and legend to match and describe the lines respectively. In our review of existing datasets, we found a mix of both private and public datasets focused on graphics digitization. However, despite their similar goal, these datasets often lack diversity in terms of plot variations, richness, and classes.
\subsubsection{Split} The \textit{LG} dataset consists of three subsets - training, validation, and test - each of which is split with a reasonable proportion of the total instances. As depicted in Table \ref{instances_split}, some classes exhibit limited numbers, however, this accurately reflects their low-frequency occurrence, such as plot titles that are typically found in figure captions. Despite this, our experiments have demonstrated that the richness of the data was crucial in overcoming this challenge. 

\begin{table}[t]
\centering
\caption{The number of semantic instances of each data split.}
\label{instances_split}
\begin{tabular}{l|c|r|r|r|r|r|r|r|r|r|r} \toprule
\multirow{2}{*}{\textbf{Split}} & \multirow{2}{*}{\textbf{\#images}} &\multicolumn{3}{c|}{\textbf{Title}} & \multicolumn{3}{c|}{\textbf{Spine}} & \multicolumn{2}{c|}{\textbf{Label}} & \textbf{Legend} & \textbf{Line}  \\ 
\cline{3-12}
& & \textit{ptitle} & \textit{xtitle} & \textit{ytitle} & \textit{xspine} & \textit{yspine} & \textit{spine} & \textit{xlabel} & \textit{ylabel} & \textit{legend} & \textit{line} \\ \midrule  \hline
Train & 312 & 39 & 268 & 235 & 245 & 255 & 431 & 1484 & 1512 & 382 & 581 \\ \hline
Validation & 104 & 11 & 92 & 78 & 82 & 82 & 163 & 539 & 484 & 152 & 183 \\ \hline
Test & 104 & 14 & 91 & 70 & 79 & 78 & 145 & 478 & 459 & 161 & 171 \\ \bottomrule
\end{tabular}
\end{table}

\begin{table}[t]
\centering
\caption{Variations of line chart categories found in \emph{LG} dataset.}
\label{variation_table}
\begin{tabular}{l|l} \hline
\textbf{Class} & \textbf{Variations} \\ \hline
Plot & combination chart, multi-axis chart, gridlines  \\ 
Title, Label & orientation, type (integer, date, etc..), font variations   \\ 
Line, Legend & style, markers, annotations, position, count \\ 
Spine & length, width, count, style  \\ 
Background & colour (gradient, filled, etc..), grid \\ 
\hline
\end{tabular}
\end{table}
\subsubsection{Variations}
Table \ref{variation_table} below demonstrates the diversity and inclusiveness of our dataset, as it includes a wide range of instances counts, styles, and locations, without any aforementioned limitations. Our dataset includes a comprehensive range of variations for all classes, as summarized in Table \ref{variation_table}. We have covered a wide range of plot types, including those that feature multiple chart types like bar, scatter, and line charts as in Figure \ref{fig1} (a) (c) and (d), as well as plots with repeated classes like multiple y or x-axes and ticks. The text content in our dataset is annotated with variations in integer, decimal, and DateTime formats, as well as tilt. Furthermore, we have taken into account different markers, patterns, and sizes for line and spine classes, and added the class "other" to represent the annotated plot area explanatory text, focus points, and arrows. The background variations in our dataset include colour (single or multiple), gradient, and RGB images.

\subsubsection{Spatial Distribution Visualization}
\begin{figure}[t]
\centering
\includegraphics[width=\textwidth]{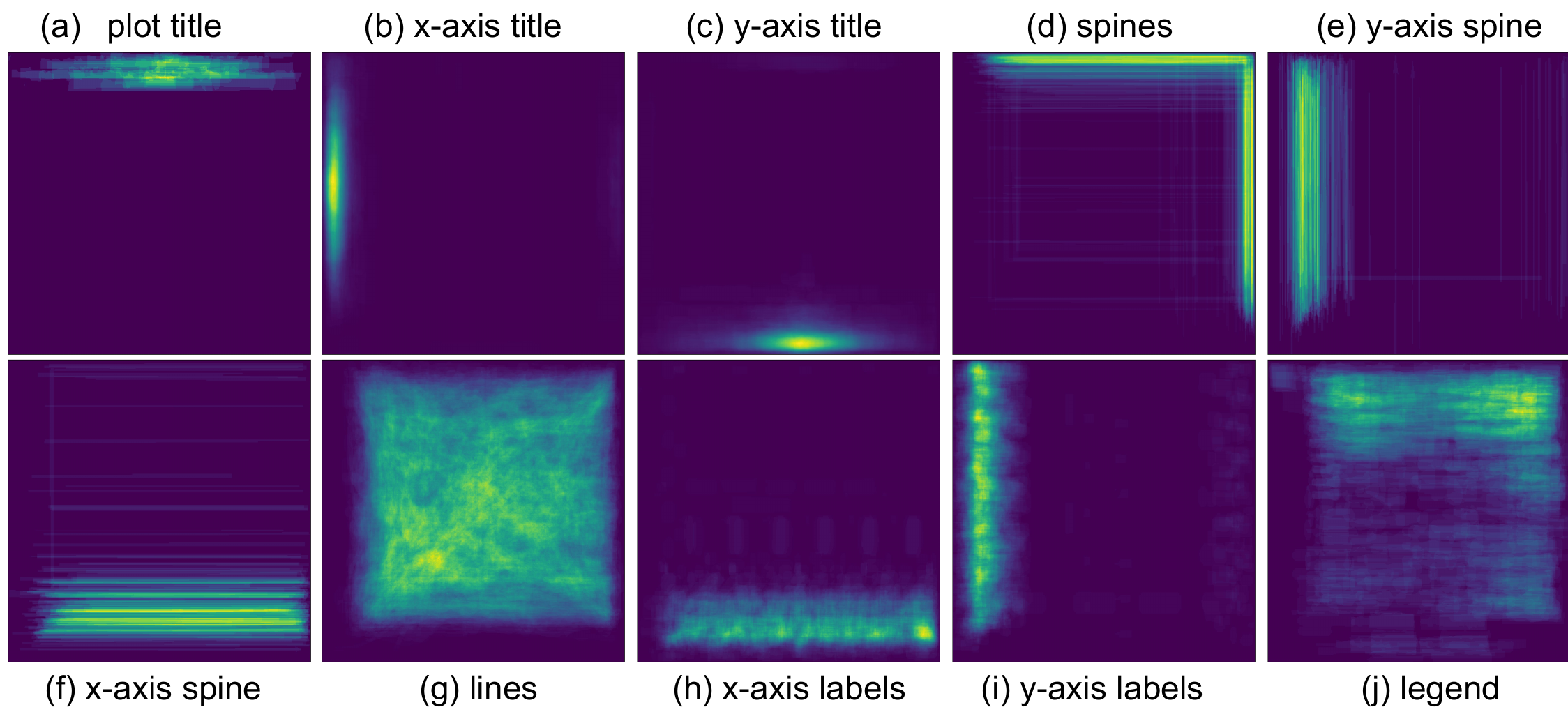}
\caption{Location heat maps for the ten different fine-grained labels in our LG dataset. We observe clear category-specific biases, (e.g., plot title mostly appears at the top, while lines are nearly uniformly distributed along the document space), which is expected for real document data.} 
\label{local_heat_maps}
\end{figure}

We have additionally analyzed the statistical localization information of all ground-truth instances. As shown in Figure \ref{local_heat_maps}, the frequency of occurrence is visualized in heatmaps. We can see that the Title class has a strong prior position, as titles are typically standalone text at the edges of the chart. Spines on the other side reveal that the majority of charts are box format with two axes (left and bottom edge). Cartesian-type with intersecting x and y axes are observed less frequently. Our heatmap evaluation shows that spines have an average width of 3 pixels, with a minimum and maximum of 1 and 7 pixels, respectively, making them one of the challenging classes to segment. Interestingly, as we see in the legend heat map, they are predominantly positioned at the top and on either side of the plotting area, but they can also appear in other locations. According to statistics, 44\% of the legends are located at the top.

%
\section{Experiments}
\subsection{Implementation Details}
We perform experiments utilizing both Jittor and Pytorch. Our implementation is based on the MMsegmentation\footnote{\url{https://github.com/open-mmlab/mmsegmentation}} library and the models were trained on an A40 GPU with an input resolution of (2048, 1024). Our evaluation metric is Mean Intersection over Union (mIoU). During training, we applied common data augmentation techniques such as random flipping, scaling (ranging from 0.5 to 2), and cropping. The batch size was set to 8 with an initial learning rate of $6e{-}5$, using a poly-learning rate decay policy. The models were trained for 50K iterations. For testing, we employed a single-scale 
flip strategy to ensure fairness in comparison.
To understand the choice of models, we further analyze the properties of the selected models in conjunction with our proposed line graphic segmentation task.

\subsection{Baselines} 

We consider 7 state-of-the-art semantic segmentation models for this task:

\textbf{MobileNetV3}~\cite{howard2019searching} is designed for image segmentation in both high and low-resource environments. It incorporates the depthwise separable convolution from MobileNetV1 and the Inverted Residual with Linear Disability from MobileNetV2 to balance accuracy and computational cost. Additionally, the model introduces the V3 lightweight attention mechanism, which enhances its ability to selectively focus on important features. These improvements make MobileNetV3 a good choice for resource-constrained applications of line graphic segmentation. 

\textbf{HRNet}~\cite{wang2021hrnet} leverages multi-scale feature representations to effectively handle high-resolution image understanding tasks such as human pose estimation and semantic segmentation. Throughout the network, HRNet utilizes repeated information exchange across multiple scales and a multi-scale feature output that is fed into a task-specific head. This innovative architecture enables easy feature reuse and efficient computation, making HRNet a top choice for high-resolution graphic segmentation task. 

\textbf{DeepLabv3+}~\cite{chen2018deeplabv3+} is a CNN semantic segmentation model that utilizes a decoder module to obtain sharper object boundaries and a more fine-grained segmentation, which is crucial for the proposed line graphic segmentation. In addition, an ASPP module captures multi-scale context information, while a lightweight Xception architecture provides efficient computation.

\textbf{PSPNet}~\cite{zhao2017pspnet} proposed the pyramid pooling module, which is able to extract an effective global contextual prior. Extracting pyramidal features since the model can perceive more context information of the input line graphic.

\textbf{Swin}~\cite{liu2021swin} introduced the Transformer model with a shifted window operation and a hierarchical design. The shifted window operation includes non-overlapping local windows and overlapping cross-windows. It provides the locality of convolution to the graphic segmentation task, and on the other hand, it saves computation as compared to the original transformer models. 

\textbf{SegFormer}~\cite{xie2021segformer} leverages the Transformer architecture and self-attention mechanisms. The model consists of a hierarchical Transformer encoder and a lightweight all-MLP decoder head, making it both effective and computationally efficient. The design enables SegFormer to capture long-range dependencies within an image, \textit{i.e.,} a line graphic in this work.

\textbf{SegNeXt}~\cite{guo2022segnext} proposed a new design of convolutional attention by rethinking the self-attention in Transformer models. In this work, the resource-costly self-attention module is replaced by using depth-wise convolution with large sizes. As a result, the multi-scale convolutional attention with a large kernel can encode context information more effectively and efficiently, which is crucial for the line graphic segmentation task.

\section{Evaluation}
\subsection{Quantitative Results}

In Table~\ref{tab3}, the efficiency-oriented CNN model MobileNetV3 with only $1.14$M parameters obtains $56.22\%$ mIoU score on the proposed LG dataset. The high-resolution model HRNet has $57.60\%$ in mIoU and the DeepLabv3+ model has $61.64\%$, but both have parameter ${>}60$M. We found that the PSPNet with pyramid pooling module in the architectural design can achieve better results with $62.04\%$. In Table~\ref{tab3}, the recent Transformer-based models achieve relatively better results than the CNN-based models. For example, the SegFormer model with pyramid architecture and with $81.97$M parameters can obtain $65.59\%$ in mIoU with a ${+}3.55\%$ gain compared to PSPNet. The Swin Transformer with hierarchical design and shifted windows has $66.61\%$ in mIoU, but it has the highest number of parameters. However, the state-of-the-art CNN-based SegNeXt utilizes multi-scale convolutional attention to evoke spatial attention, leading to the highest mIoU score of $67.56\%$ in our LG dataset. Furthermore, the SegNeXt achieve 4 top scores on 5 coarse classes, including \textit{Title, Spine, Label} and \textit{Legend}. Besides, it obtains 6 top scores out of 10 fine classes, which are \textit{xtitle, ytitle, xpsine, yspine, xlabel,} and \textit{legend}. The results show that a stronger architecture for the semantic segmentation task can achieve better results in the proposed LG benchmark, yielding reliable and accessible mathematical graphics.



\begin{table}[!t]
\caption{Semantic segmentation results of CNN- and Transformer-based models on the \emph{test} set of LG dataset. \textbf{\#P}: the number of model parameters in millions; \textbf{GFLOPs}: the model complexity calculated in the same image resolution of $512{\times}512$; \textbf{Per-class IoU} (\%): the Intersection over Union~(IoU) score for each of coarse and fine classes; \textbf{mIoU} (\%): the average score across all of $10$ fine classes. The best score is highlighted in bold.}\label{tab3}

\resizebox{\textwidth}{!}{
\begin{tabular}{l|l|cc|l|l|l|l|l|l} \toprule
\multirow{2}{*}{\textbf{Model}} & \multirow{2}{*}{\textbf{Backbone}} &\multirow{2}{*}{\textbf{\#P(M)}} & \multirow{2}{*}{\textbf{GFLOPs}} & \multicolumn{5}{c|}{\textbf{Coarse Per-class IoU}} & \multirow{2}{*}{\textbf{mIoU}} \\ \cline{5-9}
& & & & \textit{Title} & \textit{Spine} & \textit{Label} & \textit{Legend} & \textit{Line} & \\ \midrule \midrule
MobileNetV3~\cite{howard2019searching} & MobileNetV3-D8 & 1.14 & 4.20 & 45.06 & 43.68& 68.74& 60.86& 62.12& 56.22 \\ 
HRNet~\cite{wang2021hrnet} & HRNet-W48 & 65.86 & 93.59& 52.48& 44.4& 67.95& 53.34& 61.91& 57.60 \\ 
DeepLabv3+~\cite{chen2018deeplabv3+} & ResNet-50 & 62.58 & 79.15& 55.41& 46.14 & 74.72& 67.07& 32.97& 61.46 \\ 
PSPNet~\cite{zhao2017pspnet} & ResNetV1c & 144.07 & 393.90& 57.12& 43.77& 78.52& 67.75& 62.30& 62.04 \\ 
SegFormer~\cite{xie2021segformer} & MiT-B5 & 81.97 & 51.90& 58.36& 54.13& 76.79& 67.09& {69.67}& 65.59 \\ 
Swin~\cite{liu2021swin} & Swin-L & 233.65 & 403.78& 62.26& 52.85& 76.57& 68.91& \textbf{71.01}& 66.61 \\ 
SegNeXt~\cite{guo2022segnext} & MSCAN-L & 49.00 & 570.0& \textbf{63.79}& \textbf{54.61}& \textbf{80.29}& \textbf{69.09}& 65.07&\textbf{67.56} \\ \bottomrule


\end{tabular}
}
\vskip 2ex
\resizebox{\textwidth}{!}{
\begin{tabular}{l|l|l|l|l|l|l|l|l|l|l} \toprule
\multirow{2}{*}{\textbf{Model}} & \multicolumn{10}{c}{\textbf{Fine Per-class IoU}} \\ \cline{2-11}
&\textit{ptitle} & \textit{xtitle} & \textit{ytitle} & \textit{xspine} & \textit{yspine} & \textit{spine} & \textit{xlabel} & \textit{ylabel} & \textit{legend} & \textit{line} \\ \midrule \midrule
MobileNetV3~\cite{howard2019searching} & 09.03 & 55.36 & 70.81 & 53.47 & 40.21 & 37.36 & 67.83 & 69.65 & 60.86 & 62.12  \\ 
HRNet~\cite{wang2021hrnet} & 31.30 & 55.85 & 70.30 & 44.68 & 50.20 & 38.36 & 65.42 & 70.48 & 53.34 & 61.91\\ 
DeepLabv3+~\cite{chen2018deeplabv3+} & 30.30 & 62.21 & 73.74 & 49.47 & 47.57 & 41.39 & 73.65 & 75.79 & 67.07 & 32.97 \\ 
PSPNet~\cite{zhao2017pspnet} & 22.92 & 68.09 & 80.39 & 47.09 & 53.11 & 31.12 & 77.28 & 79.76 & 67.75 & 62.30  \\ 
SegFormer~\cite{xie2021segformer} & 37.25 & 61.10 & 76.75 & 60.37 & 55.83 & \textbf{46.21} & 73.35 & 80.23 & 67.09 & 69.67 \\ 
Swin~\cite{liu2021swin} & \textbf{49.21} & 59.44 & 78.15 & 59.48 & 55.33 & 43.74 & 71.96 & \textbf{81.54} & 68.91 & \textbf{71.01}  \\ 
SegNeXt~\cite{guo2022segnext} & 36.57 & \textbf{73.95} & \textbf{80.85} & \textbf{61.77} & \textbf{56.20} & 45.88 & \textbf{80.68} & 79.91 & \textbf{69.09} & 65.07 \\ \bottomrule


\end{tabular}}
\vskip -2ex
\end{table}

\subsection{Ablation Study}
Apart from the qualitative analysis of state-of-the-art models, we further perform an ablation study on the aforementioned CNN- and Transformer-based models. As shown in Table~\ref{tab4}, the ablation study is two-fold. First, to understand the impact of model scales on segmentation performance, different model scales are ablated. For example, the tiny (T) and small (S) versions of Swin Transformer, are evaluated on our LG dataset. Besides, to analyze the effect of the image resolution, two image sizes (\textit{i.e.}, $512{\times}512$ and $2048{\times}1024$) are involved in the ablation study. According to the results shown in Table~\ref{tab4}, we obtain an insight that higher resolution of input images can achieve larger gains than using models with higher complexity. For example, SegFormer-B0 in resolution of $2048{\times}1024$ outperforms SegFormer-B5 in $512{\times}512$ with a $21.82\%$ gain in mIoU. Regarding the ablation study, it is recommended to use a larger image size as input than to use a model with greater complexity. Another benefit of this setting is to maintain the high efficiency of the trained model, which is more practical and promising on graphic segmentation applications.

\begin{table}
\centering
\caption{Ablation on different models, backbone variants, and input sizes}\label{tab4}
\begin{tabular}{l|l|c|l} \hline

\textbf{Model} & \textbf{Backbone} & \textbf{Input Size} & \textbf{mIoU}\\ \hline
DeepLabv3+~\cite{chen2018deeplabv3+} & ResNet-50 &  \multirow{5}{*}{$512{\times}512$} & 22.88 \\ 
Swin~\cite{liu2021swin} & Swin-S & & 42.39\\ 
SegFormer~\cite{xie2021segformer} & MiT-B0 && 36.45 \\ 
SegNext~\cite{guo2022segnext} & MSCA-S & & \textbf{48.84} \\ 
\hline
DeepLabv3+~\cite{chen2018deeplabv3+} & ResNet-101  & \multirow{4}{*}{$512{\times}512$} & 29.11 \\
Swin~\cite{liu2021swin} &Swin-T & & 45.78 \\ 
SegFormer~\cite{xie2021segformer} & MiT-B5 & & 38.32 \\ 
SegNext~\cite{guo2022segnext} & MSCA-L & & \textbf{48.65} \\
\hline
DeepLabv3+~\cite{chen2018deeplabv3+} & ResNet-50  &\multirow{5}{*}{$2048{\times}1024$} & 63.69 \\ 
Swin~\cite{liu2021swin}& Swin-S & & 65.25 \\ 
SegFormer~\cite{xie2021segformer} & MiT-B0 &  & 60.14 \\ 
SegNext~\cite{guo2022segnext} & MSCA-S & & \textbf{67.08} \\ 
\hline

\end{tabular}
\end{table}

\begin{figure}[!t]
\centering
\includegraphics[width=\textwidth]{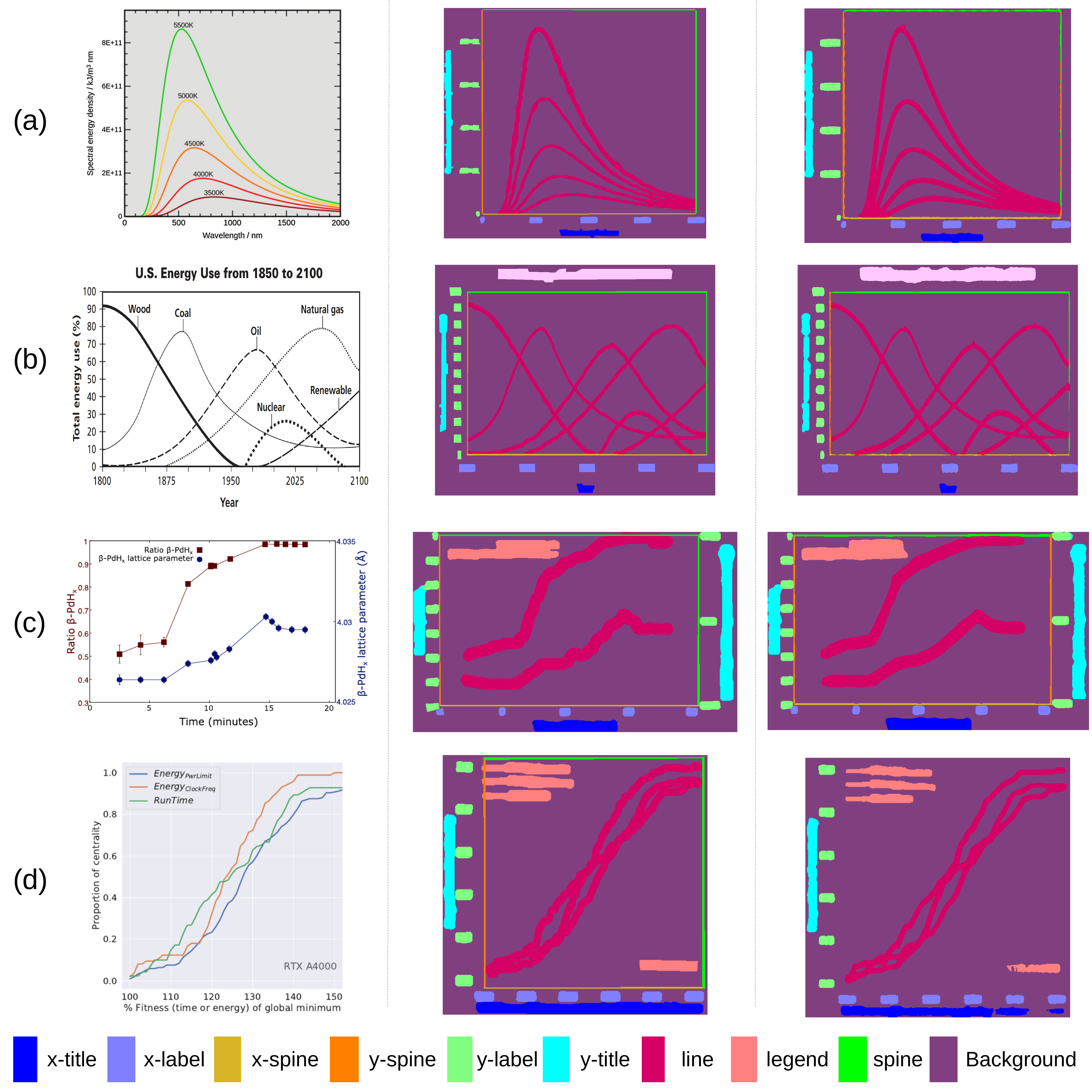}
\caption{ Visualization of semantic segmentation results of LG dataset. From left to right are the input RGB line graphs, the ground truth labels, and the predicted masks from SegNeXt-L model.
} 
\label{fig3}
\end{figure}

\subsection{Qualitative Results}

To further understand the line graph semantic segmentation task, we visualize some examples in Fig.~\ref{fig3}. From left to right are the input RGB line graphs, the ground truth labels, and the segmentation results generated by SegNeXt-L model. Although the backgrounds of these input images have different colors and textures, the model can accurately segment them (in purple). We found that SegNeXt with ${>}67\%$ mIoU can output surprisingly good segmentation results, including precise masks for thin objects, such as \textit{xspine} and \textit{yspine}. Besides, in the bottom row, the intersecting lines can be segmented accurately. Apart from the positive results, the \textit{xspine} and \textit{yspine} in the bottom row cannot be recognized well, which means that there is still room for improvement in the LG benchmark. Nonetheless, the other classes, such as \textit{labels}, \textit{titles} and \textit{legend}, can be segmented correctly. 


%
\section{Conclusion}
In conclusion, this paper presents the first line plot dataset for multi-task deep learning, providing support for object detection, semantic segmentation, and instance-level segmentation. Our comprehensive evaluations of state-of-the-art segmentation models demonstrate the potential for an end-to-end solution.
Moreover, this work has the potential to greatly improve accessibility for visually impaired and blind individuals. The ability to accurately detect and recognize mathematical graphics could lead to more accessible educational materials and support the digitization of mathematical information.
We are actively working to expand the scope of the dataset by including more types of mathematical graphics and incorporating instances relationship into the metadata. This will continue to drive advancements in this field and enable further research into the digitization of mathematical graphics.

\noindent\textbf{Acknowledgments.}
This work was supported in part by the European Union’s Horizon $2020$ research and innovation program under the Marie Sklodowska-Curie Grant No.$861166$, in part by the Ministry of Science, Research and the Arts of Baden-Württemberg (MWK) through the Cooperative Graduate School Accessibility through AI-based Assistive Technology (KATE) under Grant BW6-03, and in part by the Federal Ministry of Education and Research (BMBF) through a fellowship within the IFI programme of the German Academic Exchange Service (DAAD). This work was partially performed on the HoreKa supercomputer funded by the MWK and by the Federal Ministry of Education and Research.

%
%
\bibliographystyle{splncs04}
\bibliography{thebibliography} 

\begin{thebibliography}{10}
\providecommand{\url}[1]{\texttt{#1}}
\providecommand{\urlprefix}{URL }
\providecommand{\doi}[1]{https://doi.org/#1}

\bibitem{amin2001page}
Amin, A., Shiu, R.: Page segmentation and classification utilizing bottom-up
  approach. International Journal of Image and Graphics  \textbf{1}(02),
  345--361 (2001)

\bibitem{bajic2022multi}
Baji{\'c}, F., Orel, O., Habijan, M.: A multi-purpose shallow convolutional
  neural network for chart images. Sensors  \textbf{22}(20), ~7695 (2022)

\bibitem{breuel2017robust}
Breuel, T.M.: Robust, simple page segmentation using hybrid convolutional
  mdlstm networks. In: 2017 14th IAPR international conference on document
  analysis and recognition (ICDAR). vol.~1, pp. 733--740. IEEE (2017)

\bibitem{chen2017convolutional}
Chen, K., Seuret, M., Hennebert, J., Ingold, R.: Convolutional neural networks
  for page segmentation of historical document images. In: 2017 14th IAPR
  International Conference on Document Analysis and Recognition (ICDAR).
  vol.~1, pp. 965--970. IEEE (2017)

\bibitem{chen2015page}
Chen, K., Seuret, M., Liwicki, M., Hennebert, J., Ingold, R.: Page segmentation
  of historical document images with convolutional autoencoders. In: 2015 13th
  International Conference on Document Analysis and Recognition (ICDAR). pp.
  1011--1015. IEEE (2015)

\bibitem{chen2018deeplabv3+}
Chen, L.C., Zhu, Y., Papandreou, G., Schroff, F., Adam, H.: Encoder-decoder
  with atrous separable convolution for semantic image segmentation. In: ECCV
  (2018)

\bibitem{chintalapati2022dataset}
Chintalapati, S., Bragg, J., Wang, L.L.: A dataset of alt texts from hci
  publications: Analyses and uses towards producing more descriptive alt texts
  of data visualizations in scientific papers. arXiv preprint arXiv:2209.13718
  (2022)

\bibitem{choi2019visualizing}
Choi, J., Jung, S., Park, D.G., Choo, J., Elmqvist, N.: Visualizing for the
  non-visual: Enabling the visually impaired to use visualization. In: Computer
  Graphics Forum. pp. 249--260. Wiley Online Library (2019)

\bibitem{clark2016pdffigures}
Clark, C., Divvala, S.: Pdffigures 2.0: Mining figures from research papers.
  In: Proceedings of the 16th ACM/IEEE-CS on Joint Conference on Digital
  Libraries. pp. 143--152 (2016)

\bibitem{clausner2017icdar2017}
Clausner, C., Antonacopoulos, A., Pletschacher, S.: Icdar2017 competition on
  recognition of documents with complex layouts-rdcl2017. In: 2017 14th IAPR
  International Conference on Document Analysis and Recognition (ICDAR).
  vol.~1, pp. 1404--1410. IEEE (2017)

\bibitem{dai2018chart}
Dai, W., Wang, M., Niu, Z., Zhang, J.: Chart decoder: Generating textual and
  numeric information from chart images automatically. Journal of Visual
  Languages \& Computing  \textbf{48},  101--109 (2018)

\bibitem{davila2019icdar}
Davila, K., Kota, B.U., Setlur, S., Govindaraju, V., Tensmeyer, C., Shekhar,
  S., Chaudhry, R.: Icdar 2019 competition on harvesting raw tables from
  infographics (chart-infographics). In: 2019 International Conference on
  Document Analysis and Recognition (ICDAR). pp. 1594--1599. IEEE (2019)

\bibitem{davila2020chart}
Davila, K., Setlur, S., Doermann, D., Kota, B.U., Govindaraju, V.: Chart
  mining: A survey of methods for automated chart analysis. IEEE transactions
  on pattern analysis and machine intelligence  \textbf{43}(11),  3799--3819
  (2020)

\bibitem{davila2021icpr}
Davila, K., Tensmeyer, C., Shekhar, S., Singh, H., Setlur, S., Govindaraju, V.:
  Icpr 2020-competition on harvesting raw tables from infographics. In: Pattern
  Recognition. ICPR International Workshops and Challenges: Virtual Event,
  January 10-15, 2021, Proceedings, Part VIII. pp. 361--380. Springer (2021)

\bibitem{drivas1995page}
Drivas, D., Amin, A.: Page segmentation and classification utilising a
  bottom-up approach. In: Proceedings of 3rd International Conference on
  Document Analysis and Recognition. vol.~2, pp. 610--614. IEEE (1995)

\bibitem{guo2022segnext}
Guo, M.H., Lu, C.Z., Hou, Q., Liu, Z., Cheng, M.M., Hu, S.M.: Segnext:
  Rethinking convolutional attention design for semantic segmentation. arXiv
  preprint arXiv:2209.08575  (2022)

\bibitem{ha1995document}
Ha, J., Haralick, R.M., Phillips, I.T.: Document page decomposition by the
  bounding-box project. In: Proceedings of 3rd International Conference on
  Document Analysis and Recognition. vol.~2, pp. 1119--1122. IEEE (1995)

\bibitem{haurilet2019spase}
Haurilet, M., Al-Halah, Z., Stiefelhagen, R.: Spase-multi-label page
  segmentation for presentation slides. In: 2019 IEEE Winter Conference on
  Applications of Computer Vision (WACV). pp. 726--734. IEEE (2019)

\bibitem{haurilet2019wise}
Haurilet, M., Roitberg, A., Martinez, M., Stiefelhagen, R.: Wise—slide
  segmentation in the wild. In: 2019 International Conference on Document
  Analysis and Recognition (ICDAR). pp. 343--348. IEEE (2019)

\bibitem{howard2019searching}
Howard, A., Sandler, M., Chu, G., Chen, L.C., Chen, B., Tan, M., Wang, W., Zhu,
  Y., Pang, R., Vasudevan, V., et~al.: Searching for mobilenetv3. In:
  Proceedings of the IEEE/CVF international conference on computer vision. pp.
  1314--1324 (2019)

\bibitem{huang2019icdar2019}
Huang, Z., Chen, K., He, J., Bai, X., Karatzas, D., Lu, S., Jawahar, C.:
  Icdar2019 competition on scanned receipt ocr and information extraction. In:
  2019 International Conference on Document Analysis and Recognition (ICDAR).
  pp. 1516--1520. IEEE (2019)

\bibitem{jobin2019docfigure}
Jobin, K., Mondal, A., Jawahar, C.: Docfigure: A dataset for scientific
  document figure classification. In: 2019 International Conference on Document
  Analysis and Recognition Workshops (ICDARW). vol.~1, pp. 74--79. IEEE (2019)

\bibitem{keefer2014image}
Keefer, R., Bourbakis, N.: From image to xml: Monitoring a page layout analysis
  approach for the visually impaired. International Journal of Monitoring and
  Surveillance Technologies Research (IJMSTR)  \textbf{2}(1),  22--43 (2014)

\bibitem{li2019figure}
Li, P., Jiang, X., Shatkay, H.: Figure and caption extraction from biomedical
  documents. Bioinformatics  \textbf{35}(21),  4381--4388 (2019)

\bibitem{liu2019data}
Liu, X., Klabjan, D., NBless, P.: Data extraction from charts via single deep
  neural network. arXiv preprint arXiv:1906.11906  (2019)

\bibitem{liu2021swin}
Liu, Z., Lin, Y., Cao, Y., Hu, H., Wei, Y., Zhang, Z., Lin, S., Guo, B.: Swin
  transformer: Hierarchical vision transformer using shifted windows. In: ICCV
  (2021)

\bibitem{long2018textsnake}
Long, S., Ruan, J., Zhang, W., He, X., Wu, W., Yao, C.: Textsnake: A flexible
  representation for detecting text of arbitrary shapes. In: Proceedings of the
  European conference on computer vision (ECCV). pp. 20--36 (2018)

\bibitem{methani2020plotqa}
Methani, N., Ganguly, P., Khapra, M.M., Kumar, P.: Plotqa: Reasoning over
  scientific plots. In: Proceedings of the IEEE/CVF Winter Conference on
  Applications of Computer Vision. pp. 1527--1536 (2020)

\bibitem{poco2017reverse}
Poco, J., Heer, J.: Reverse-engineering visualizations: Recovering visual
  encodings from chart images. In: Computer graphics forum. pp. 353--363. Wiley
  Online Library (2017)

\bibitem{seweryn2021will}
Seweryn, K., Lorenc, K., Wr{\'o}blewska, A., Sysko-Roma{\'n}czuk, S.: What will
  you tell me about the chart?--automated description of charts. In:
  International Conference on Neural Information Processing. pp. 12--19.
  Springer (2021)

\bibitem{siegel2018extracting}
Siegel, N., Lourie, N., Power, R., Ammar, W.: Extracting scientific figures
  with distantly supervised neural networks. In: Proceedings of the 18th
  ACM/IEEE on joint conference on digital libraries. pp. 223--232 (2018)

\bibitem{wang2021hrnet}
Wang, J., Sun, K., Cheng, T., Jiang, B., Deng, C., Zhao, Y., Liu, D., Mu, Y.,
  Tan, M., Wang, X., Liu, W., Xiao, B.: Deep high-resolution representation
  learning for visual recognition. TPAMI  (2021)

\bibitem{xie2021segformer}
Xie, E., Wang, W., Yu, Z., Anandkumar, A., Alvarez, J.M., Luo, P.: {SegFormer:}
  {Simple} and efficient design for semantic segmentation with transformers.
  In: NeurIPS (2021)

\bibitem{yang2017learning}
Yang, X., Yumer, E., Asente, P., Kraley, M., Kifer, D., Lee~Giles, C.: Learning
  to extract semantic structure from documents using multimodal fully
  convolutional neural networks. In: Proceedings of the IEEE Conference on
  Computer Vision and Pattern Recognition. pp. 5315--5324 (2017)

\bibitem{yoshitake2020program}
YOSHITAKE, M., KONO, T., KADOHIRA, T.: Program for automatic numerical
  conversion of a line graph (line plot). Journal of Computer Chemistry, Japan
  \textbf{19}(2),  25--35 (2020)

\bibitem{zhang2021transfer}
Zhang, J., Ma, C., Yang, K., Roitberg, A., Peng, K., Stiefelhagen, R.: Transfer
  beyond the field of view: Dense panoramic semantic segmentation via
  unsupervised domain adaptation. IEEE Transactions on Intelligent
  Transportation Systems  \textbf{23}(7),  9478--9491 (2021)

\bibitem{zhao2017pspnet}
Zhao, H., Shi, J., Qi, X., Wang, X., Jia, J.: Pyramid scene parsing network.
  In: CVPR (2017)

\end{thebibliography}
\end{document}